\documentclass[preprint,12pt]{elsarticle}

\usepackage{graphicx}
\usepackage{amssymb}

\usepackage{lineno}
\usepackage{url}

\journal{arXiv}

\begin{document}

\begin{frontmatter}


\title{Encouraging an Appropriate Representation Simplifies Training of Neural Networks}



\author{Krisztian Buza$^{a,b}$}

\address{$^{a}$ Department of Artificial Intelligence, Faculty of Informatics, \\ E\"ov\"os Lor\'and Unviersity (ELTE), Budapest, Hungary, buza@biointelligence.hu \\
$^{b}$ Center for the Study of Complexity, Babe\c s-Bolyai University, Cluj Napoca, Romania}

\begin{abstract}

A common assumption about neural networks is that they can learn an appropriate internal representations on their own, see e.g. end-to-end learning. In this work we challenge this assumption. We consider two simple tasks and show that the state-of-the-art training algorithm fails, although the model itself is able to represent an appropriate solution. We will demonstrate that encouraging an appropriate internal representation allows the same model to solve these tasks. While we do not claim that it is impossible to solve these tasks by other means (such as neural networks with more layers), our results illustrate that integration of domain knowledge in form of a desired internal representation may improve the generalisation ability of neural networks.  
\end{abstract}

\begin{keyword}
Neural networks \sep Representation \sep Integration of domain knowledge


\end{keyword}

\end{frontmatter}



\section{Introduction}

Traditionally, the applications of machine learning algorithms were closely coupled with careful feature engineering requiring extensive domain knowledge. In contrast, in the era of deep learning, a common assumption is that neural networks are able to learn appropriate representations that may be better than the features defined by domain experts, see end-to-end learning for self-driving cars~\cite{bojarski2016end}, speech recognition~\cite{amodei2016deep} and other applications~\cite{gordo2017end, silver2017predictron}.

Recent success stories related to neural networks include mastering board games~\cite{silver2016mastering}, the diagnosis of various diseases, such as skin cancer~\cite{esteva2017dermatologist}, retinal disease~\cite{de2018clinically} and mild cognitive impairment~\cite{meszlenyi2017resting}. Despite these (and many other) spectacular results, there is increasing evidence indicating that neural networks do not learn the underlying concepts: minor alterations of images, that are invisible to humans, may lead to erroneous recognition~\cite{szegedy2013intriguing, papernot2017practical}, e.g., slight modifications of traffic signs ``can completely fool machine learning algorithms"~\cite{ackerman2019slight}.

In this paper we will demonstrate that encouraging an appropriate representation may substantially simplify the training of (deep) neural networks. In particular, we consider two simple tasks and show that the state-of-the-art training algorithm fails, although the model itself is able to represent an appropriate solution. Subsequently, we ill see that encouraging an appropriate internal representation allows the same model to solve the same tasks. The resulting networks not only have good generalization abilities, but they are more understandable to human experts as well. 

The reminder of the paper is organised as follows: next, we will explain what we mean by ``encouraging a representation" (Section~\ref{sec:enc}). In Section~\ref{sec:tranet} and Section~\ref{sec:results} we will demonstrate in the context of two transformation tasks that the proposed idea may substantially improve the accuracy of the model. Finally, we discuss the implications of our observations to other applications in Section~\ref{sec:discussion}.  

\section{Encouraging a representation}
\label{sec:enc}

With ``encouraging a representation" we mean to train a neural network in a way that some of the hidden units correspond to predefined concepts. We would like to emphasize that this requirement is meant for a relatively small subset of all the hidden units. For example, if we want to train a network for the recognition of traffic signs in images, the network is likely to have thousands of hidden units and we may require that some of those hidden nodes recognize particular shapes or letters, i.e., their activations should be related to the presence of a triangle, an octagon, or particular letters like 'S', 'T', 'O' and 'P'.   

The idea of encouraging a representation may be implemented in various ways: for example, in the cost function, we may include a regularisation term that penalises the situation if the activation of some given nodes is inappropriate. In one of our previous works, in the context of matrix factorisation for drug-target interaction prediction, we encouraged a lower dimensional representation so that the distances between drugs and targets are in accordance with their chemical and genomic similarities~\cite{peska2017drug}.

In this paper, we consider encoder-decoder networks. In particular, we will ``encourage" the encoder to learn a pre-defined representation. This representations is the output of the encoder and serves as the input of the decoder. We demonstrate that encouraging an appropriate representation may lead to substantial increase of the accuracy.

\section{TraNet: a network for Translation and Transcription}
\label{sec:tranet}

We consider two tasks, Translation and Transcription (see Section~\ref{sec:tasks}). We try to solve both of these tasks with the same neural network, called \emph{TraNet} (as ``Tran" is a  common prefix of the names of these two tasks). When solving the two tasks, the only difference is the input layer in accordance with the input of the tasks. 

We implemented all the experiments in Python using numpy, matplotlib and TensorFlow with the keras API. In order to assist reproducibility, our implementation is available at
\url{http://www.biointelligence.hu/encourage/}~.

\subsection{Benchmark tasks: Translation and Transcription}
\label{sec:tasks}

\begin{figure}[t]
\centering\includegraphics[width=\columnwidth]{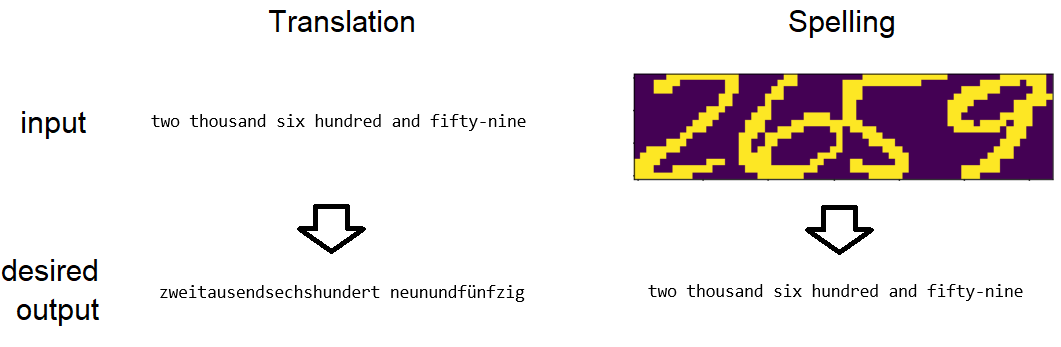}
\caption{\label{fig:tasks} Illustration of the considered tasks, Translation (left) and Transcription (right).}
\end{figure}

As benchmark tasks, we consider the translation of written numbers from English to German (e.g. \emph{twenty-five} should be translated to \emph{funfundzwanzig}\footnote{For simplicity, instead of the German special letters '\"a', '\"o', '\"u' and '$\beta$', we use 'a', 'o', 'u' and 'ss' respectively.}), and recognition of 4-digit handwritten numbers where the desired output is the number written in English, see also Fig~\ref{fig:tasks}. For simplicity, we will refer to these tasks as \emph{Translation} and \emph{Transcription}. 

As neural networks are used for substantially more complex tasks, see e.g.~\cite{zhang2015deep} for a review, we expect them to have an excellent performance in case of our tasks. 

\subsection{Architecture}

TraNet is a feed-forward encoder-decoder network. For strings (i.e. numbers written in English or German) we use letter-wise one-hot encoding. For example, 'a' is coded as $(1, 0, 0, 0, ..., 0)$, 'b' is coded as $(0, 1, 0, 0, ..., 0)$, etc. The entire sting is coded as the concatenation of the codes of each letter. We allow at most a length of 50 letters, therefore,
the output layer of TraNet contains $50 \times 29 = 1450$ units. Similarly, in case of the Translation task, the input layer of TraNet contains 1450 units as well. When TraNet is used for Transcription, the input is a binary image with $64 \times 16 = 1024$ pixels, therefore, the input layer contains 1024 units, each one corresponding to one of the pixels. The input layer does not perform any operation, its solely purpose is to represent the input data.

\label{sec:representation}
An appropriate internal representation, which \emph{could} be learned by the network, is the \emph{digit-wise one-hot encoding} of the 4-digit number that was shown to the network. That is: '0' is coded as $(1, 0, 0, 0, ..., 0)$, '1' is coded as $(0, 1, 0, 0, ..., 0)$, etc. The digit-wise one-hot encoding of a 4-digit number is the concatenation of the vectors corresponding to the digits of the number, therefore, it has a total length of $4 \times 10 = 40$. 

TraNet contains 3 hidden layers. The first and third layers contain 1000 units with ReLU activation function. The second hidden layer contains 40 units so that it might potentially learn the aforementioned digit-wise one-hot encoding. In the second layer, we use the sigmoid activation function, as it may be suited to the digit-wise one-hot encoding.     

As loss function, we use binary cross-entropy. We trained TraNet for 100 epochs using the ADAM optimizer~\cite{kingma2014adam}. We performed the computations on CPU (i.e., no GPU/TPU support was used).

\section{Results}
\label{sec:results}

Next, we compare conventional training and training with encouraging the digit-wise one-hot encoding as internal representation.

\subsection{Translation}

In case of Translation, we considered the numbers between 0 and 9999, written in English (input) and German (desired output). A randomly selected subset of 100 numbers were used as test data, while the remaining 9900 instances were used as training data. We repeated the experiments 5-times with different initial split of the data into training and test data.

In case of conventional training, denoted as \emph{Conventional TraNet}, the resulting network was not able to give an exact translation for any of the numbers of the test set. While in some cases the translations generated by the network were at least partially understandable to humans, in other cases the network failed to translate the numbers, see Tab~\ref{tab:translation} for some examples. 

Although there are many possibilities to improve the model, next, we show that there is nothing wrong with the model: TraNet itself is able to represent a function that gives a reasonably good translation, the problem is the conventional training. 

In order to encourage the model to learn an appropriate representation, we train the encoder and decoder separately. The encoder consist of the input layer and the first two hidden layers. It is trained to translate from English to the digit-wise one-hot encoding (described in Section~\ref{sec:representation}), i.e., we expect the output of the encoder to be the digit-wise one-hot encoding. 
The input of the decoder is the second hidden layer of TraNet. Additionally, the third hidden layer of TraNet and its output layer belong to the decoder. The decoder is trained to translate from digit-wise one-hot encoding to German. 

When training the encoder and decoder separately, so that the the digit-wise one-hot encoding is encouraged, we observed that the network was able to translate on average 95.8 \% of the numbers perfectly. From the practical point of view, the quality of the translation is even better, because in many cases when the translation was not perfect, we observed only minor spelling mistakes, such as ``einhundert einundneenzig" instead of ``einhundert einundneunzig", or ``siebenhundert sechsundsehhzig" instead of ``siebenhundert sechsundsechzig". Consequently, encouraging an appropriate representation lead to an accurate model for translation of numbers from English to German.

\begin{table}[t]
\centering
\caption{\label{tab:translation} Examples for translation of numbers from English to German \emph{with} and \emph{without} encouraging the digit-wise one-hot encoding as internal representation, denoted as ``encouraged TraNet" and ``conventional TraNet", respectively.}
\begin{tabular}{l l l}
\hline
\footnotesize \textbf{Input} & 
\footnotesize \textbf{Output of} & 
\footnotesize \textbf{Output of}\\
 & 
\footnotesize \textbf{conventional TraNet} & 
\footnotesize \textbf{encouraged TraNet}\\
\hline
\footnotesize one hundred and ninety one & 
\footnotesize einaaudenaaaiaaaaaaaaaa & 
\footnotesize einhundert einundneenzig \\
\hline
\footnotesize four thousand & 
\footnotesize vieatausenazieihundert & 
\footnotesize viertausendzweihundert  \\
\footnotesize two hundred and twenty-five & 
\footnotesize aieaanaaaanai & 
\footnotesize funfundzwanzig \\
\hline
\footnotesize eight thousand &
\footnotesize acattausenaaaeihundert &
\footnotesize achttausendachthundert \\
\footnotesize eight hundred and sixty &
\footnotesize aieaaaaa &
\footnotesize sechzig \\
\hline
\footnotesize seven hundred and sixty-six &
\footnotesize aaebenaunaeaaaaaanan- &
\footnotesize siebenhundert sechsundsehhzig \\
&
\footnotesize aaaaaiaeaa &
\\
\hline
\normalsize
\end{tabular}

\end{table}

\subsection{Transcription}

In order to obtain 4-digit handwritten numbers, we considered the \emph{Semeion} dataset\footnote{\url{https://archive.ics.uci.edu/ml/datasets/Semeion+Handwritten+Digit}}, which is a publicly available dataset containing images of handwritten digits. As each of the images shows a single digit, in order to obtain an image of a four-digit handwritten number, we choose four images randomly and stack them horizontally, see Fig.~\ref{fig:tasks} for an example. The last 100 images of the Semeion dataset are used to obtain test images, whereas the training data is obtained from the first 1493 images. In total, we obtained 100~000 training images and 1000 test images of 4-digit numbers. We repeated the experiment 5-times with different training and test images.

\begin{figure}[t]
\centering\includegraphics[width=\columnwidth]{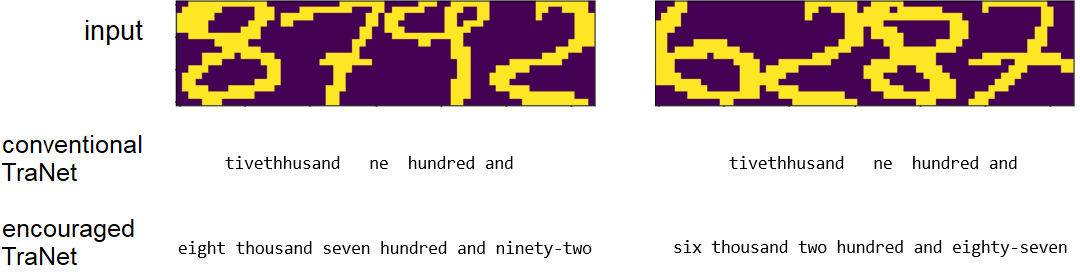}
\caption{\label{fig:transcription} Examples for transcription \emph{with} and \emph{without} encouraging the digit-wise one-hot encoding as internal representation, denoted as ``encouraged TraNet" (bottom) and ``conventional TraNet" (center), respectively.}
\end{figure}

Conventional training of TraNet, i.e., training without encouraging the digit-wise one-hot encoding as intermediate representation, lead to a network that could not transcribe any of the images correctly. Instead, the output of TraNet was always a number-like phrase without clear meaning, such as ``fivet huusadd ne  hundded and" or ``tivethhusand ne hundred and". In contrast, encouraging the digit-wise one-hot encoding, i.e., training the encoder and decoder separately, lead to a network that was able to perfectly transcribe on average 74.9 \% numbers of the test data, while it made mostly minor spelling mistakes in the remaining cases, see Fig.~\ref{fig:transcription} for some illustrative examples.

\section{Discussion}
\label{sec:discussion}

While we do not claim that there are no other ways to find an acceptable solution, in fact, we observed that using more deep neural networks achieved comparable performance to that of encouraged TraNet\footnote{We note that, due to its simplicity, our translation task can be solved perfectly using techniques from the classic theory of formal languages.}, our results clearly demonstrate the power of encouraging the digit-wise one-hot encoding.  

Besides allowing a relatively simple model to achieve good generalisation ability, our encouraged models are better understandable to domain experts. On the one hand, this may increase the trust in the model, and it allows to ``debug" the model: if the output is incorrect, one could check whether the activations in the 2nd hidden layer are close to that of in case of the desired internal representation. 

Although the fact that a neural network may learn concepts that are substantially different from human concepts, may be considered as an advantage in many cases, we argue that it may be worth to use encouraged models together with conventional models. For example, in critical applications (credit scoring, medical diagnosis, etc.), in may be worth to carefully consider cases when conventional and encouraged models disagree.  

Although the appropriate representation depends on the underlying task, we believe that the general idea of encouraging models to learn concepts that are similar to human concepts may be beneficial in many applications, especially in cases when minor modifications of the input cause ``completely fool"~\cite{ackerman2019slight} the learning algorithm. Defining an appropriate representation may be seen as a way of integrating domain knowledge into the training procedure, thus it can be seen as a way of collaboration between human intelligence and machine intelligence.

\section*{Acknowledgements}

\noindent
K. Buza was supported by Thematic Excellence Programme, Industry and Digitization Subprogramme, NRDI Office, 2019 and received the ``Professor Ferencz Rad\'o"  Fellowship of the Babes-Bolyai University, Cluj Napoca, Romania. 






\end{document}